\pdfoutput=1

\documentclass[11pt]{article}
\usepackage{tabularx}
\usepackage{graphicx}
\graphicspath{ {./image/} }

\usepackage[]{ACL2023}

\usepackage{lipsum}

\usepackage{times}
\usepackage{latexsym}

\usepackage[T1]{fontenc}

\usepackage[utf8]{inputenc}
\usepackage{inconsolata}

\title{Extraction and Summarization of Explicit Video Content using Multi-Modal Deep Learning}

\author{Shaunak Joshi\thanks{\ \ These authors contributed equally to this work.}\textsuperscript{*†} \hspace{1cm} Raghav Gaggar\footnotemark[1]\textsuperscript{*†} \\
  {\tt shaunaks@usc.edu} \hspace{1cm} {\tt gaggar@usc.edu} \\
  University of Southern California\textsuperscript{\dag}}

\begin{document}
\maketitle
\begin{abstract}
With the increase in video-sharing platforms across the internet, it is difficult for humans to moderate the data for explicit content. Hence, an automated pipeline to scan through video data for explicit content has become the need of the hour. We propose a novel pipeline that uses multi-modal deep learning to first extract the explicit segments of input videos and then summarize their content using text to determine its age appropriateness and age rating. We also evaluate our pipeline's effectiveness in the end using standard metrics. \\
\end{abstract}

\section{Introduction}
Explicit/violent material can potentially harm the psychological well-being of viewers. Historically, such material in videos has been controlled by censorship or regulation. In popular culture, movies or television programs are often required to be rated by a system like the MPA rating system, which consists of human verification of videos. The goal of these systems was to inform audiences of the age appropriateness of the video so that the viewers have pertinent information beforehand about the content. However, as online video-sharing social media platforms such as YouTube and TikTok gained popularity, uncontrolled video materials became more widespread, increasing potential harm to general audiences across the internet. Therefore, given the scale of video content being generated, it is essential that its scanning and summarizing are done automatically.

In this paper, we propose a novel pipeline to fully automate the process of classification and summarization of explicit/violent segments of videos using multi-modal deep-learning techniques. In the context of this paper, we exclusively focus on explicit content containing information of a violent nature. Each segment within a video will be split into an array of image frames, an audio segment, and a text segment. It will then be processed through their respective deep learning models for classification and summary of segments.

Analysis of contents will check for \textit{violent} features. "Violent content" refers to the visual and audio depiction of brutal, realistic acts of violence.

One major challenge in detecting and summarizing such content in videos is the sheer volume of data available online. With millions of videos uploaded every day, it's simply not feasible for humans to review each one. This is where machine learning techniques can be particularly helpful, as they will help by automating certain parts of the pipeline.

Another challenge is ensuring that the algorithms used to scan and summarize videos are accurate and reliable. It's also important to consider potential biases in the data, such as over- or under-representation of certain types of content or demographics. Hence, we need to choose and train the machine learning models carefully.

In addition to the technical challenges, there are also ethical and legal considerations that must be taken into account when developing such systems. For example, there is a risk of false positives or false negatives, which could result in inappropriate censorship or the dissemination of harmful content. There are also concerns around privacy and data protection, particularly if the systems involve collecting and analyzing personal information about video viewers.

Despite these challenges, there is a growing need for automated systems to detect and summarize explicit or violent content in videos, particularly on social media platforms where such content can be easily shared and viewed by large audiences.

In this research paper, we discuss our pipeline to detect and summarize explicit or violent content in videos. We also discuss the approaches and settings we have used, the risks and challenges we have faced, and finally, our results. We hope that our work will be used to develop an effective pipeline for automated explicit content scanning and summarization, which will ultimately lead to a safer and more informed online viewing experience. Also, we believe our work has the potential to be expanded for utilization in real-world applications such as movie, broadcast, and video game censorship.\\

\section{Related Work}
Explicit or violent material could potentially harm the psychological well-being, especially of young viewers. Previous studies have indicated that such videos can have negative effects on young viewers, such as increased aggression and anxiety \cite{wilson2008media, chang2019effect} and can also raise the likelihood of them engaging in risky behaviors such as alcohol and drug consumption, or can lead them to unwanted pregnancies \cite{STRASBURGER1989747}. 

Various rating systems, such as the Motion Picture Association of America (MPAA) rating system, have been used to inform viewers of the age appropriateness of video content. The video gaming industry also has established its own Entertainment Software Rating Board (ESRB) regulation to assign age and content ratings to video games in the United States and Canada. However, these systems are often subjective and rely on human intervention, which can lead to inconsistencies in rating. MPAA rating has shown its limitation of not being able to control the exposure to drug-related content in movies \cite{tickle2009tobacco}. Meanwhile, as the popularity of short-form video providers such as TikTok or Youtube Shorts rose, the burden on such organizations has increased. Therefore, the need for a more automated and objective system for rating explicit or violent content has become crucial.

The use of deep learning models has shown promising results in various tasks, including image recognition, speech recognition, and natural language processing. By leveraging deep learning, we can automatically identify explicit or violent content from different modalities, including image frames, audio, and text, thereby providing a more comprehensive and objective analysis.

\cite{yousaf2022deep} Fed the sequence of extracted video frames to their CNN-LSTM-FNN-based pipeline to classify the frames into safe, fantasy violence, or sexual nudity. The approach used by \cite{sharma22_interspeech} included feeding audio files to a CNN architecture to detect whether a given input includes violence or not. \cite{zhang2021none} Used movie text scripts to identify the severity of age-restricted content. They used transformers for encoding the texts, which were then fed to an LSTM-based deep learning model. These approaches utilized single modality information from input videos. But with recent advancements in computing, we can now extract features from multiple modalities that are embedded in a video, simultaneously.

Multi-modal deep learning involves utilizing different modalities to learn the representations of a single input. The models based on them are generally better at capturing information and learning representations as compared to models utilizing uni-modal information.

Hence, recent research on this topic involved employing multi-modal approaches to detect explicit or violent videos. One study on harmful content detection in videos \cite{edstedt2022vidharm} used SlowFast \cite{Feichtenhofer_2019_ICCV} networks for modeling the visual data and ResNet \cite{he2015deep} model for the audio data. Then the outputs from both of these models were fused using an MLP to give the final prediction. In \cite{shafaei2021case}, the researchers took raw frames, dialogue subtitles, and raw audio from videos and fed each of them to separate deep learning pipelines to learn their vector representations. The vectors obtained were then fused using Gated Multi-modal Unit (GMU), Late Fusion, and Feature Concatenation Fusion to detect whether the input movie trailer was appropriate for children or not. 

The previous studies failed to address further explainability for the final classification i.e. if a video is classified as negative, exactly why it is classified as negative. Also, only movie trailers were explored in this domain, and no work has been done on longer duration videos.

We address the shortcomings of the previous works in the following manner:
\begin{enumerate}
\item Our work will be capable of taking a long video and approximately localizing its segments which contain explicit content.
\item  For each video segment that is identified as explicit, we provide its natural language summarization which mainly answers the question of why the segment has been classified as explicit. The summary that is generated will explain what is happening in the explicit video segment, which will ultimately result in adding a major component of explainability that the previous researchers failed to address.\\
\end{enumerate}

\section{Methodology}

\begin{figure*}[h]
    \centering
    \includegraphics[width=16cm,height= 8cm]{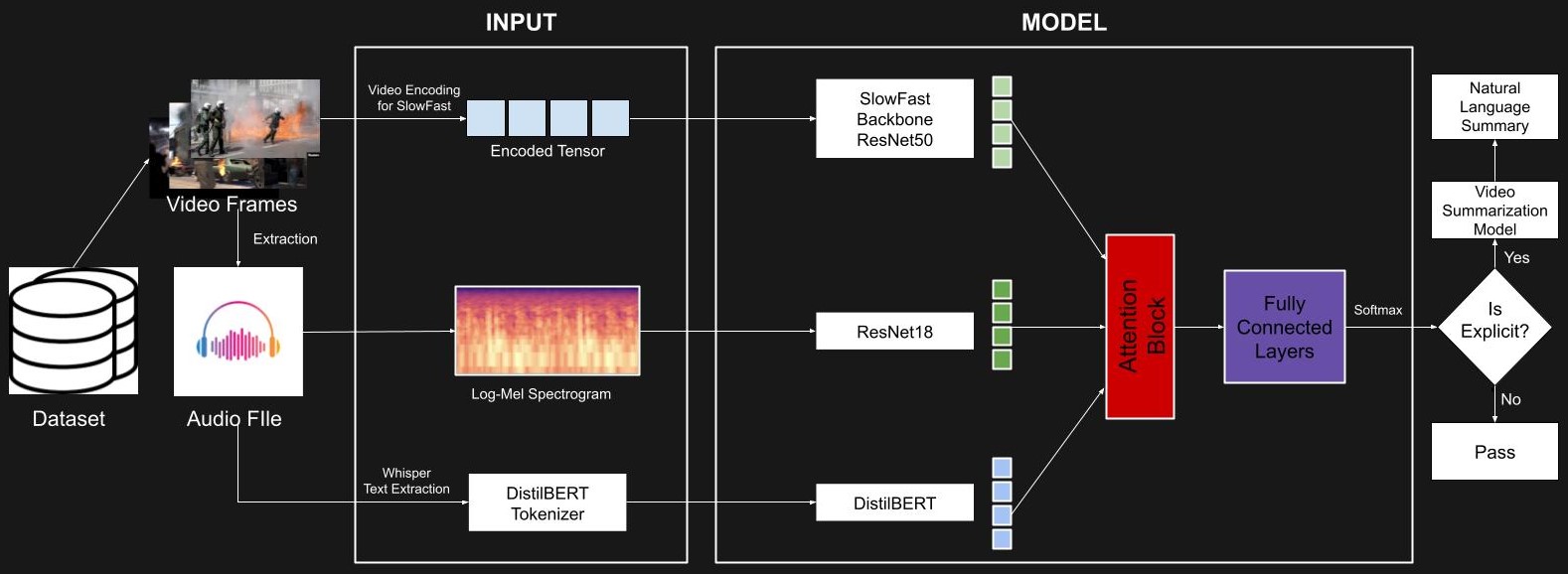}
    \caption{Pipeline}
    \label{fig:my_label1}
\end{figure*}

\begin{figure*}[h]
    \centering
    \includegraphics[width=5.5cm,height= 7cm]{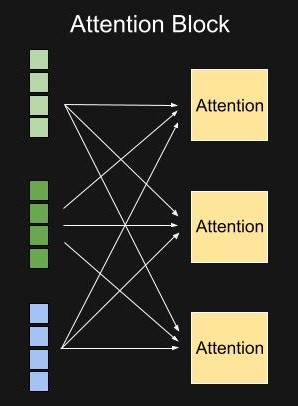}
    \caption{Attention Mechanism}
    \label{fig:my_label2}
\end{figure*}

Our pipeline, as depicted in Figure \ref{fig:my_label1} can be divided into 2 parts. First, we break down the videos into 1-minute video clips that are fed sequentially to our classifier. Next, if the segment is classified as explicit, it is fed to a video summarizer model for generating its natural language summary. 

For our dataset we have borrowed a subset from  \textit{XD-Violence} \cite{wu2020look} for training and testing our models. This section describes the methodology in detail that we have used to achieve our goal.

\subsection{Video Segment Classifier}
In the first part, we will initially clip the raw videos into parts having a duration of 1 minute or less. Not every video within the chosen datasets was feasible to use as input for our models. Some videos did not have the audio necessary for extracting audio and textual features. After clipping of videos into segments, we split each segment's information into its three modalities - visual, auditory, and written. Data from each modality is then be fed to separate deep-learning models.

\subsubsection{Video Modality}
To capture the information stored in video frames, we will encode the video frames into a 4-dimensional tensor having the size of (num\_frames, image\_channels, height, width). We used SlowFast \cite{feichtenhofer2019slowfast} model having ResNet-50 \cite{He_2016_CVPR} as the backbone, which serves as the feature extractor for video. SlowFast contains two pathways - the slow path and the fast path. The slow path takes all the consecutive frames of the video in order to capture the spatial semantics whereas the fast path skips frames at uniform intervals to just capture motion. We encode the above 4-dimensional tensors into two lists of tensors, one list for each path in the SlowFast model. We chose the SlowFast model for getting video features because of its state-of-the-art performance on the Kinetics \cite{kay2017kinetics} dataset. This dataset is predominantly used for action classification, and a model pre-trained for this task would be a good choice as a base for our downstream task of determining whether an action performed in the clip is explicit or not.

\subsubsection{Audio Modality}
The second part of our video segment classifier works with audio data. Our goal is to provide additional information about explicit or violent content through the information stored in the audio of the video clips. We used the mel-spectrograms of audio data and fed them to a ResNet-18 model for feature extraction. ResNet-18 is chosen because of its higher performance and lower computational requirements as compared to similar image recognition architectures like VGG \cite{simonyan2015deep}. \\

\subsubsection{Language Modality}
We also extract the dialogue texts by feeding the processed input videos to OpenAI Whisper \cite{radford2022robust}, which is one of the best open-source language models that transcribes audio data into text.

Whisper feeds transcribed data to a HuggingFace DistilBERT \cite{sanh2019distilbert} model, which generates the features of the textual data. We selected DistilBERT because it is powerful and lightweight at the same time, in comparison to the rest of the state-of-the-art transformers. DistilBERT is 40 percent less in size in comparison to the original BERT model \cite{devlin2019bert}, though it retains 97 percent of its language understanding capabilities and is 60 percent faster. Since we were constrained by our computing resources, we chose DistilBERT.

\subsubsection{TriModal Fusion: Combining Video, Audio, and Language}
The fusion of video, audio, and language modalities forms the most crucial task of our approach to multimodal machine learning. This subsection describes the methodologies employed to integrate these modalities into a joint representation. In Section \ref{par:Modalities and Their Predictive Power} we dive into details of determining the predictive performance of these techniques.

\paragraph{Concatenation}
This technique serves as our baseline. In this approach, the feature vectors extracted from each modality are concatenated to form a single high-dimensional feature space. This concatenated vector is then fed into a fully connected layer, which acts as the precursor to the classification layer. This strategy was used to exploit the correlations between the different modalities not taking into account their independent nature.

\paragraph{Unified Attention}
Building upon the baseline, this method tries to refine the fusion process by incorporating attention mechanism. Post concatenation of the individual modality feature vectors, the resultant vector is employed as the query (Q), key (K), and value (V) in the self-attention module. This self-attention mechanism is designed to weigh the features beyond simple correlation as performed in the baseline. However, this method retains the concatenated structure as the basis for feature interaction, which limits the model's ability to exploit modality-specific dynamics as explained further in \ref{subsec:Explicit vs non-explicit classification}

\paragraph{Combinatorial Attention}
This approach represents a somewhat more advanced fusion method explored in our study. Changing our approach from the unified feature space, this technique utilizes each modality's feature vector in a pairwise fashion to generate three unique combinations that serve as inputs to separate attention mechanisms as depicted in Figure \ref{fig:my_label2}. Each pair's attention output is then concatenated, creating a stacked feature map that. This approach utilizes the inter-modality interactions at a finer granularity. The final feature map is subsequently processed by a fully connected layer, followed by the classification layer. This approach is hypothesized to provide a deeper understanding of the data by allowing each modality to interact with the others through the attention framework, potentially understanding complex patterns that cannot be grasped by Concatenation and Unified Attention.

\subsection{Summarization}

The goal of the second part of our pipeline is to summarize the explicit video segments using natural language through zero-shot learning. In order to achieve that, the segment classified as explicit in the first part of our pipeline would be further fed to a video summarization model. A pre-trained GIT model \cite{wang2022git} from HuggingFace was used to achieve the zero-shot English summarization task. First, each explicit segment is broken down into smaller chunks, which are fed to the pre-trained model sequentially. Breaking down explicit segments into smaller chunks allows us to capture the finer details of the complete segment, which are otherwise missed by the summarizer model as observed in our experiments. The text output of each chunk of the explicit segment is then finally concatenated to get an English language summary of the content present in the complete segment. \\

Overall, our pipeline takes a video as input, isolates its explicit segments, and then generates a natural language summary of them.\\

\section{Experiment Settings}

\subsection{Dataset}
XD-Violence dataset contains short video clips belonging to one of these 7 classes - Car Accident, Explosion, Fighting, Riot, Shooting, and Normal Activities. We programmatically tagged the videos that belonged to 6 classes apart from Normal Activities as \textit{explicit} and normal activities as \textit{non-explicit}. The videos from this dataset also contain audio information. The final dataset used by us comprised 1659 video segments. More details on our dataset are shown in Table 1 below.

\begin{center}
\begin{table}[h!]
    \centering
    \resizebox{.55\columnwidth}{!}{%
    \begin{tabular}{ c|c|c }
        Class & Train & Test \\
        \hline
        Explicit & 521 & 130 \\ 

        \hline
        Non-explicit & 807 & 201 \\

        \hline
        Total & 1328 & 331 \\

    \end{tabular}%
    }
    \caption{Number of samples in our train and test set}
    \label{table:1}
\end{table}
\vspace{-1cm}
\end{center}

\subsection{Training Details}
Our classifier is built and trained end-to-end to detect whether the input video segment is explicit/violent or not. Since our fully connected layers have a softmax function in the end, we used cross-entropy as our loss function for the binary classification of video segments. We used SGD optimizer with its default learning rate of 1e-3 and momentum of 0.9 for all our models and trained our joint model for 100 epochs. A batch size of 1 was used for this model. We used two NVIDIA 1080 Ti GPUs with 11 GB VRAM each for faster training of this joint model.

\begin{table}[]
\resizebox{\columnwidth}{!}{%
\begin{tabular}{|c|c|c|c|}
\hline
            & F1-Micro & F1-Macro & F1-Weighted \\ \hline
Concatenation       & 0.82     & 0.81     & 0.82        \\ \hline
Unified Attention   & 0.82     & 0.81     & 0.82        \\ \hline
Combinatorial Attention & \textbf{0.83} & \textbf{0.82} & \textbf{0.83} \\ \hline
\end{tabular}}
\caption{Comparison of Concatenation, Unified Attention, and Combinatorial Attention models on the validation dataset}
\label{table:2}
\end{table}

\begin{table}[]
\resizebox{\columnwidth}{!}{%
\begin{tabular}{|l|c|c|c|}
\hline
            & F1-Micro & F1-Macro & F1-Weighted \\ \hline
UniModal    & \begin{tabular}[c]{@{}c@{}} Video: 0.70 \\ Language: 0.75 \\ Audio: 0.61 \end{tabular} &  \begin{tabular}[c]{@{}c@{}} Video: 0.75 \\ Language:0.73 \\ Audio:0.38 \end{tabular} & \begin{tabular}[c]{@{}c@{}} Video:0.77 \\Language:0.74 \\Audio:0.46 \end{tabular} \\ \hline
            
BiModal     & \begin{tabular}[c]{@{}c@{}} Language and Audio:0.78 \\
Language and Video: 0.81 \\ Video and Audio:0.78 \end{tabular} &
\begin{tabular}[c]{@{}c@{}} Language and Audio:0.78 \\ Language and Video: 0.80 \\ Video and Audio:0.75 \end{tabular} &

 \begin{tabular}[c]{@{}c@{}} Language and Audio:0.78 \\
Language and Video: 0.81 \\ Video and Audio:0.77 \end{tabular} \\
\hline

TriModal    & \textbf{0.82} & \textbf{0.81} & \textbf{0.82} \\ \hline
\end{tabular}}
\caption{Comparison of unimodal, bimodal, and trimodal models for our baseline on the validation dataset}
\label{table:3}
\vspace{-0.2cm}
\end{table}

\subsection{Evaluation}
\subsubsection{Explicit vs non-explicit classification}
\label{subsec:Explicit vs non-explicit classification}
\paragraph{Cross-Modal Feature Fusion Evaluation}
   We conducted a study that measures the effectiveness of the fusion methods we used by looking at the key performance metrics: F1-Micro, F1-Macro, and F1-Weighted scores. In table \ref{table:2} the Concatenation approach provided a reliable baseline, with consistent scores across all metrics. This suggests that even a simple fusion strategy can be quite effective for multimodal feature integration. 
   
   Now in order to experiment with a better fusion technique as opposed to the baseline we trained the classifier using Unified Attention. This method did not result in any performance gains which might be due to a suboptimal use of attention to fuse the modalities in question. Concatenating the modalities and using self-attention on the resultant tensor might have reduced our control over how the distinct modalities interacted thus hindering the unique contributions of each modality to the overall representation. 

   Further observing this limitation we optimised our fusion even further by training it using Combinatorial Attention. It resulted in a better performance which could be seen as a direct response to the limitations presented by the Unified Attention approach. While the Unified Attention method's practice of fusing modalities prior to self-attention might have constrained our control over the fusion process, the Combinatorial Attention method allowed for a better control over the interaction of these modalities. According to Table \ref{table:2} comparing all the fusion approaches discussed above Combinatorial Attention outperforms the concatenation and the Unified Attention approach.

\phantomsection
\label{par:Modalities and Their Predictive Power}
\paragraph{Modalities and Their Predictive Power}
   We compared how models perform when they use different combinations of modalities — audio, language, video and even using them separately. The performance is reflected through F1 scores presented in Table \ref{table:3}.

   In the unimodal phase, we trained the classifier for each modality in isolation. This evaluation helped identify the inherent characteristic of each modality.

   The bimodal phase explored the potential of modality pairs taken two at a time. This experiment aimed to give an in-depth analysis of the interaction of different modalities. The results suggest a benefit when modalities were paired, particularly the language and video combination, which outperformed the language and audio pair across all metrics. These experiments pertaining to the bimodal interactions offer a valuable understanding of how different modalities complement one another to enhance model performance.

    Finally in the trimodal approach, represented a full integration of video, language, and audio. This model which is nothing but our baseline aimed to use the collective strengths of all modalities. The trimodal configuration showed an advantage, yielding the highest F1 scores, indicating that the information of combined modality features could provide a more complete understanding for classification tasks. The consistent improvement across all metrics compared to the bimodal results highlights the value of integrating a broader range of modalities.

\begin{figure*}[h]
    \centering
    \includegraphics[width=16cm,height= 8cm]{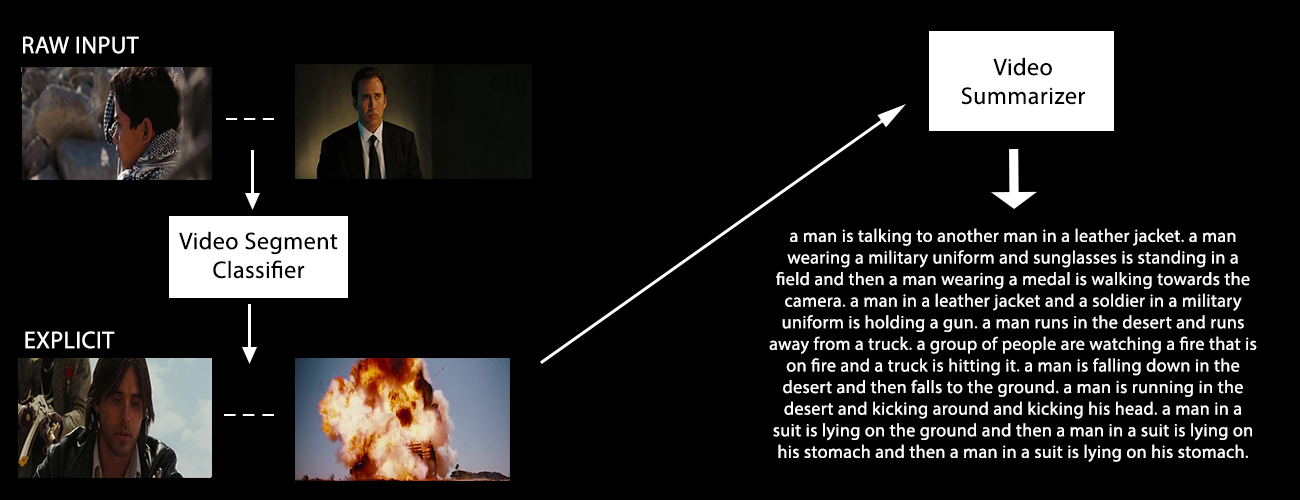}
    \caption{Results}
    \label{fig:my_label3}
\end{figure*}

\subsubsection{Video Summarization}
The segments classified using our multimodal pipeline are fed into the video summarizer model to generate a summary of explicit segments in natural language an example of which is shown in Figure \ref{fig:my_label3}. The qualitative evaluation process of the summarization allows us to assess its effectiveness in accurately generating informative summaries of those explicit input segments.\\

\section{Discussion}
We successfully employed multi-modal learning to classify whether the input video segments are explicit or violent and also generated a text summary of such segments.  

We took video clips from our XD-Violence dataset that belonged to the same movie and stitched them together to form a complete video and then ran our pipeline over it to classify the video segments containing explicit content and summarize them. In the future, sexual explicitness can also be included as part of the dataset to make it more robust. Work can also be done to train and test our pipeline on longer videos. This project also can have the capability to detect the severity of the explicit scenes in order to tag them for suitable age groups and reduce the manual efforts of corporations like IMDb and Rotten Tomatoes. 

One of the main advantages of this methodology is that it is designed to work on shorter video clips, which are more commonly found on social media platforms. Moreover, this pipeline has the potential to assist in content rating by detecting the severity of explicit scenes and tagging them for suitable age groups, not limited to a certain ethnicity or culture. Furthermore, this pipeline has the potential to reduce the manual efforts of corporations.

We lacked the resources to employ quantitative metrics for our video summarization model. Future research on this topic can explore that, and also the possibility of developing a more comprehensive rating system that can accurately detect and rate explicit and violent content in videos. Currently, our work is designed to fuse information from three modalities using attention mechanisms to provide a more objective and comprehensive analysis of video content. This can be achieved by combining deep learning models with human expertise to create a more accurate and reliable rating system.

The potential applications of this methodology extend beyond social media platforms and content rating. For example, it could be used in education to automatically filter out inappropriate content from online lectures and tutorials. It could also be utilized in law enforcement to identify explicit or violent content in videos submitted as evidence. The possibilities are extensive, and this research could be a stepping stone towards a more automated and objective approach to video content analysis.

\bibliography{custom}
\bibliographystyle{acl_natbib}

\end{document}